\pdfoutput=1

\documentclass[conference]{IEEEtran}
\IEEEoverridecommandlockouts
\usepackage{cite}
\usepackage{amsmath,amssymb,amsfonts}
\usepackage{algorithmic}
\usepackage{graphicx}
\usepackage{textcomp}
\usepackage{xcolor}
\usepackage{bm}
\usepackage{units}
\newcommand{\figref}[1]{{Fig.~\ref{#1}}}

\newcommand{\refsec}[1]{Sec. \ref{sec:#1}}
\newcommand{\secref}[1]{\refsec{#1}} 
\newcommand{\argmax}{\mathop{\rm arg~max}\limits}

\def\BibTeX{{\rm B\kern-.05em{\sc i\kern-.025em b}\kern-.08em
    T\kern-.1667em\lower.7ex\hbox{E}\kern-.125emX}}
\begin{document}

\title{  
  Circus ANYmal: A Quadruped Learning Dexterous Manipulation with Its Limbs
}

\author{Fan Shi$^{*}$,
  Timon Homberger$^{\dag}$,
  Joonho Lee$^{\dag}$,
  Takahiro Miki$^{\dag}$,
  Moju Zhao$^{*}$,
  Farbod Farshidian$^{\dag}$,\\
  Kei Okada$^{*}$,
Masayuki Inaba$^{*}$,
Marco Hutter$^{\dag}$
\\${}^{*}$ The University of Tokyo, 113-8654, Tokyo, Japan
\\${}^{\dag}$ ETH Z\"urich, 8092 Z\"urich, Switzerland
\thanks{$^{*}$
F. Shi, M. Zhao, K. Okada and M. Inaba are with JSK Lab,
Department of Creative-Infomatics, 
The University of Tokyo, 
7-3-1 Hongo, Bunkyo-ku, Tokyo 113-8656, Japan
{\tt\small shifan@jsk.t.u-tokyo.ac.jp}}
\thanks{$^{\dag}$
  T. Homberger, J. Lee, T. Miki, F. Farshidian,
M. Hutter are with
Robotics Systems Lab,
ETH Z\"urich,
LEE building, Leonhardstrasse 21, 8092 Zurich
}
\thanks{
  This research was supported by the Swiss National Science Foundation through the National Center of Competence in Research (NCCR) Robotics.
}
}

\maketitle

\bibliographystyle{IEEEtran}

\begin{abstract}
  Quadrupedal robots are skillful at locomotion tasks while lacking manipulation skills, not to mention dexterous manipulation abilities.
  Inspired by the animal behavior and the duality between multi-legged locomotion and multi-fingered manipulation, we showcase a circus ball challenge on a quadrupedal robot, ANYmal.
  We employ a model-free reinforcement learning approach to train a deep policy that enables the robot to balance and manipulate a light-weight ball robustly using its limbs without any contact measurement sensor. 
  The policy is trained in the simulation, in which we randomize many physical properties with additive noise and inject random disturbance force during manipulation, and achieves zero-shot deployment on the real robot without any adjustment.
  In the hardware experiments, dynamic performance is achieved with a maximum rotation speed of $\unit[15]{^{\circ}/s}$, and robust recovery is showcased under external poking.
  To our best knowledge, it is the first work that demonstrates the dexterous dynamic manipulation on a real quadrupedal robot.
\end{abstract}


\section{Introduction}
\label{sec:intro}
With the great progress on both hardware and algorithm, quadrupedal robots recently have shown significant performance in locomotion tasks, such as high-speed running \cite{sangbae2019mini, bdspot}, robust falling recovery \cite{hwangbo2019scirob, lee2019robust}, walking on challenging terrain \cite{tsounis2020deepgait, da2020learning, lee2020sci}.
Compared to wheeled or biped robots, thanks to the flexible limbs and larger support region, quadrupedal robots are more robust to cope with complex environments, such as stairs and uneven terrain.
Consequently, quadrupedal robots are expected to achieve various real-world missions. These vehicles have proven to be extremely robust and can cope with complex environments and terrains, which opened the potential for application in real world missions \cite{Bellicoso2018jfr, jpl2020subt}.

However, most quadrupedal robots' applications focus on navigation and inspection, while active interaction and manipulation of the environment are still lacking.
To overcome these limitations and extend the advanced locomotion capability with manipulation skills, several groups equipped their quadrupedal robots with a robotic arm and gripper\cite{iit2016quad_plus_arm, bdspot, rsl2019alma, sethu2020anymal_arm}.
This enables basic manipulation tasks such as pick-and-place, door opening, and cooperative carrying, whereby the manipulation problem is largely decoupled from locomotion.
On the downside, payload limitations of the existing quadrupedal robots mostly only allow carrying a single arm, which limits the manipulation skills.

\begin{figure}[t]
  \begin{center}
    \includegraphics[clip,  bb= 30 0 930 520,  width=1.0\columnwidth]{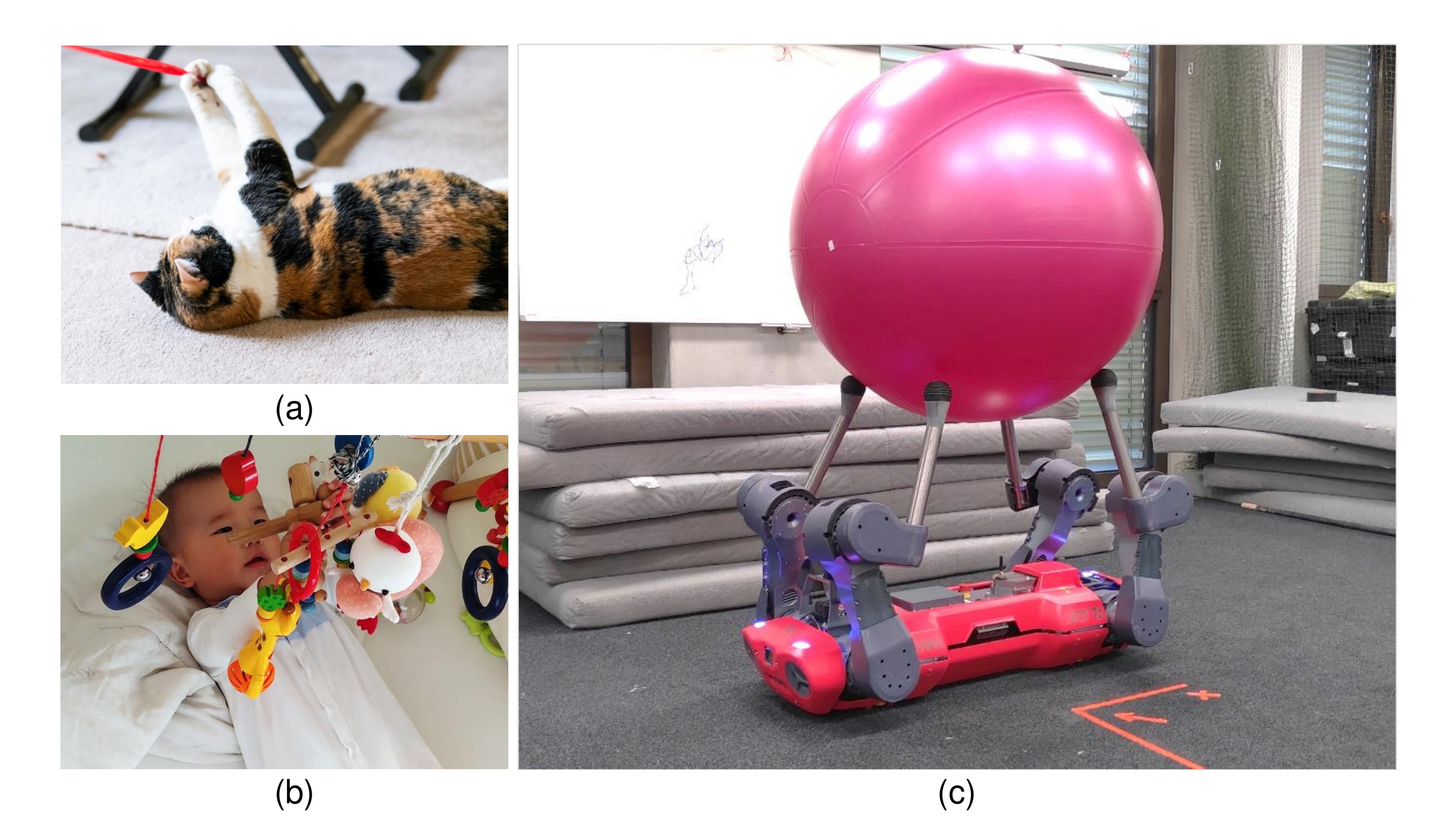}
    \caption{Inspired by the manipulation skills of animals (a) and babies (b), we enable the quadrupedal robot ANYmal (c) to manipulate objects with its legs.
      Experiment video: \textbf{https://youtu.be/lmWSw\_Hl9l8}}
    \label{figure:abstract}
  \end{center}
\end{figure}

In contrast to these developments in robotics, where locomotion and manipulation are tackled with separate hardware components, we can observe truly amazing manipulation skills in nature. Quadrupedal animals can utilize their legs to dexterously manipulate the environment (Fig.~\ref{figure:abstract}a).
Inspired by this ability, a few approaches have suggested utilizing the legs as manipulators to achieve dexterity with minimal hardware modification.
In \cite{mori_2002_whole_quad_manip, hong2020transport}, two limbs are used to pick up a box and the other two limbs for stance balancing.
For multi-legs robots such as hexapod robots, four limbs are used balancing while the other two limbs are manipulating objects \cite{noriho_1995_hexapod_manip, noriho_2000_hexapod_rise, adachi_2002_hexapod_manip, inoue2010pushing}.
Another solution is to add additional support legs to the quadrupedal robot to maximize the robustness of loco-manipulation \cite{Vijayakumar2020anymal_manip}.
Compared to dual-limbs manipulation, the most extreme idea is to synthesize all the limbs for manipulation.
Inchworm-like motion with four legs is achieved in \cite{ma2020tro} to manipulate two boards.
In \cite{ucb2020quad_ball}, a quadrupedal robot in the simulation balances on a ball and simultaneously manipulates it.

Locomotion and manipulation are regarded as the dual problem in some aspects \cite{raibert1986legged, lynch1996nonprehensile, johnson2013selfmanip}, whereby legs could be viewed as the fingers of in-hand manipulation.
Dexterous in-hand manipulation has a long history and is still a challenging problem \cite{aiyama_pivot, okamura2000dexterous_survey, bicchi2000survey, mordatch2012dexterous_sim, dollar2012dexterous, karen2014dexterous_sim, kumar2016optimal_dexterous}.
A high-DoF hand's controller must handle complex contact dynamics that are difficult to model accurately and challenging to compute online.
Recently, deep reinforcement learning shows great progress on in-hand dexterous manipulation \cite{van2015learning, vikash2017learning, barth2018dexterous_sim, plappert2018dexterous_sim, vikash2019dexterous, openai2019rubik, openai2020inhand, vikash2020inhand_ball}.
Robots can learn robust policies in simulation or real environment to operate valve \cite{vikash2019dexterous}, rotate block \cite{openai2020inhand} and even solve the Rubik's Cube \cite{openai2019rubik} in the real world.

On the other hand, for quadrupedal locomotion, deep reinforcement learning also plays a vital role in recent progress. 
Walking policy is learned on a small-size quadrupedal robot \cite{tan2018sim, haarnoja2018learning, yang2020data, ha2020learning}.
Actuator network is developed to reduce the sim-to-real gap and achieve robust dynamic locomotion skills on ANYmal robot \cite{hwangbo2019scirob, lee2019robust, lee2020sci}.
Gait selection is learned to cope with rough terrain with less energy usage \cite{da2020learning}.
Animal behavior is imitated and further improved by RL to deploy on the real robot \cite{peng2020dog}.

\subsection{Contribution}
\label{sec:intro-contribution}
Motivated by the recent progress in learning quadrupedal locomotion and in-hand manipulation, we revisit the leg-manipulation task to attain four-limbs manipulation skills. Our main contribution is to develop a framework that enables a quadrupedal robot to achieve a first step towards dexterous full-limbs manipulation.
We demonstrate our framework's effectiveness by showcasing it on a series of challenging circus tasks such as rotating a ball in roll, pitch, and yaw direction with different velocity.
To the best of our knowledge, it is the first work to achieve the quadrupedal dexterous dynamic object manipulation on a real robot.

\section{Method}
\label{sec:method}

\begin{figure}[t]
  \begin{center}
    \includegraphics[clip,  bb= 120 0 820 268,  width=1.0\columnwidth]{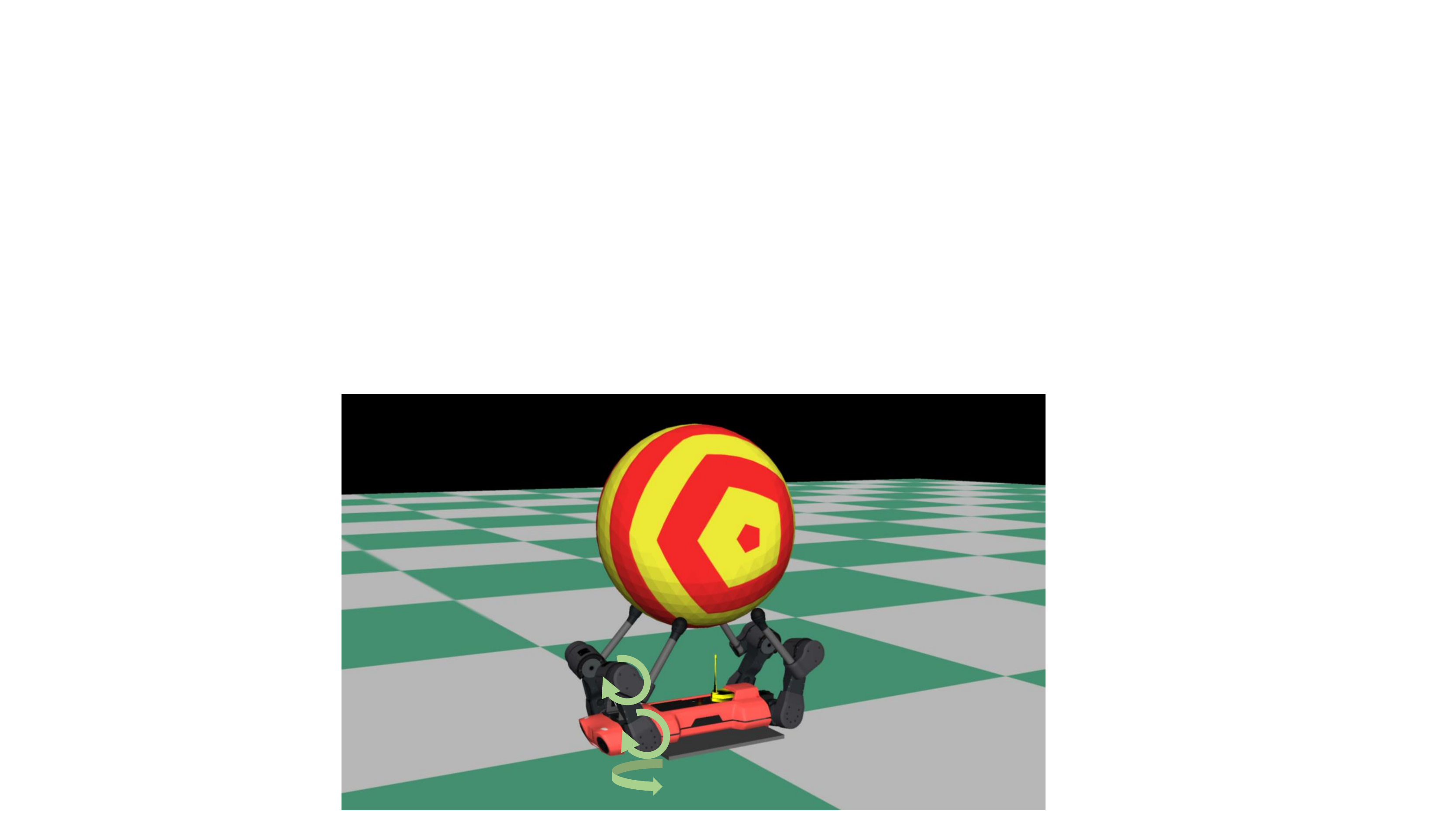}
    \caption{Circus experiment with ANYmal-C robot in the Raisim simulator \cite{hwangbo2018raisim}. Each leg has $3$ DoF driven by SEA actuators, which are Hip Abduction/Adduction (HAA), Hip Flexion/Extension (HFE), and Knee Flexion/Extension (KFE) from the bottom to the top.}
    \label{figure:model}
  \end{center}
\end{figure}

In this paper, we propose a method for in-limb manipulation with ANYmal \cite{hutter2017anymal}, a quadruped robot actuated with $12$ series elastic actuators (SEAs), as illustrated in \figref{figure:model}.
Our method allows dynamically rotating a circus ball based on user-defined target velocities.
To verify the $SO(3)$ rotation ability, we train the policy to rotate the ball in roll, pitch, and yaw directions. 
Moreover, we limit the manipulation contact points only on feet, where the robot's base should not touch the ball.

The proprioceptive measurements are limited only to the joints' position and velocity without any contact measurement such as contact state or contact force.
To further simplify the problem, position and orientation of the ball are obtained from an external motion capture system,
and mass and radius of the ball are known to the controller.
We train the policy with model-free RL in the Raisim simulator \cite{hwangbo2018raisim}, and achieve the zero-shot deployment on the real robot.

In the rest of this section, the whole learning process would be introduced, including the environment's Markov Decision Process(MDP) formulation, detailed training settings, network architecture, and policy gradient algorithm.

\subsection{Observation and Action}
\label{sec:method-obs}
The observation includes joint states, previous action, ball states, and task-related inputs.
Compared to the locomotion task \cite{hwangbo2019scirob, peng2020dog}, the position and velocity of robot base are eliminated since the robot lies on its back during the manipulation task.
The joint state includes positions and velocities measured by joint encoders on the robot.
The action in the previous time step (i.e., joint position command) is recorded and included as well.
The ball's position and orientation (relative to the robot base) is observed by the external motion capture system.
The original command is the ball's target angular velocity. 
We translate it into two parts: (1) target orientation represented by quaternion and (2) time to reach the goal orientation.
Both parts are updated periodically.

We stack the observations of the last three time steps~\cite{hwangbo2019scirob, peng2020dog}, which is the same history length as the input to the actuator model.
This enables the policy to handle latencies and partially observable states of the hardware~\cite{haarnoja2018learning}

The history of joint position $\bm\theta_{t}$, joint velocity $\dot{\bm\theta}_{t}$, action $\bm a_{t}$, ball position $\bm p_{t}$ and quaternion difference to target orientation $\bm q_{t}$ are sampled at $10$ and $20$ $ms$ previously to avoid being too sparse or dense.
Thus the observation, $\bm s_t \in \mathbb{R}^{130}$, is defined as
\begin{align}
 \bm x_t =& (
 \bm\theta_{t}, \dot{\bm\theta}_{t}, \bm p_{t}, \bm q_{t}, \bm a_{t}
 )
 \notag \\
 \bm s_t =& (\bm x_t, \bm x_{t-1}, \bm x_{t-2}, t_{remain}) . 
\end{align}

The action (output of the policy) is sent to the robot as the joint target position.
A low-level joint position PD controller translates the policy output to motor torque command.

\subsection{Reward Function}
\label{sec:method-reward}
The goal of our reward design is to encourage the robot to manipulate the ball to track the commanded speed while being intuitive and simple with a minimum number of terms.
In contrast to locomotion tasks \cite{hwangbo2019scirob, peng2020dog}, here, the speed is implicitly represented by the periodically updated target orientation.
The reasons are twofold: first, assuming sphere manipulation as the complex-terrain locomotion, it is difficult to keep constant speed~\cite{lee2020sci};
and second, for most manipulation tasks, it is not necessary to maintain constant speed, but accurate tracking for target orientation is more critical \cite{openai2020inhand, openai2019rubik}.
Thus the reward is represented as the angle difference $\delta q$ between the current and target orientation
%
%
%
\begin{align}
    r_{t}^{q} = k_q * \frac{1}{e^{\delta q} + 2 + e^{-\delta q}},
\end{align}
where $\delta q$ could be calculated from quaternion difference in the observation state: $\delta q = 2 \hspace{1mm} cos^{-1}(\bm q_{t}(0))$.

Aggressive leg motions can result in base motion, which is dangerous to the hardware and can cause task failure. To avoid this behavior and keep the base still, a negative reward is added to punish any base velocity.
\begin{equation}
  \label{eq:reward_basevel}
  r_{t}^{v} = -k_v * ||\bm v_{base}(t)||,
\end{equation}
here $\bm v_{base}(t)$ denotes the linear velocity of robot base, which could be obtained in the simulation.
This reward is dominant in the early training stage and reduces to zero in the middle stage.

Joint torque is punished for keeping the feasible and energy-efficient torque distribution and preventing high contact force during manipulation. High contact forces often denote that the legs are forcefully squeezing the ball, which can deform the object and leads to task failure.
\begin{equation}
  \label{eq:reward_torque}
  r_{t}^{\tau} = -k_{\tau} * ||\bm \tau(t)||^{2}.
\end{equation}

Although less slippage at contact points is desirable, completely avoiding it is not realistic.
It would cause a significant reality gap between simulation and real robot, which is also the limitation in the previous model-based method \cite{ucb2020quad_ball}. 
Thus, we only moderately punish the slipping velocity instead
\begin{equation}
  \label{eq:reward_slip}
  r_{t}^{slip} = 
  \begin{cases}
    -k_{slip} * ||\bm v_{\tan}(t)||, \hspace{10mm}$in contact$
    \cr
    \hspace{13mm}0, \hspace{22mm}$no contact$
  \end{cases}
\end{equation}
where $\bm v_{\tan}(t)$ denotes the relative tangent velocity between the ball and the end-effector of each leg.

Compared to walking on the stiff ground, most of the manipulated daily object is more deformable under large contact force.
Since our simulator \cite{hwangbo2018raisim} could not simulate deformation, it would lead to huge difference between the sim and real especially in the normal direction of contact forces.
In our real experiment, a yoga ball is used, which is even more deformable.
To tackle this problem, we punish the normal contact velocity.
\begin{equation}
  \label{eq:reward_deform}
  r_{t}^{collide} = 
  \begin{cases}
    -k_{collide} * ||\bm v_{norm}(t)||, \hspace{10mm}$in contact$
    \cr
    \hspace{16mm}0, \hspace{25mm}$no contact$
  \end{cases}
\end{equation}
where $\bm v_{norm}(t)$ is the relative contact velocity in the normal direction between ball and the end-effector of each leg.

Compared to the reward design in quadrupedal locomotion task \cite{hwangbo2018raisim, lee2019robust, peng2020dog, lee2020sci}, the joint speed reward, and foot clearance reward are eliminated.
The former one is implicitly contained in the robot base and contact velocity reward.
The latter one, whose essential part is the foot clearance threshold, is not intuitive to select in our setup with shaped objects compared to flat terrain.

\subsection{Early Termination}
\label{sec:method-terminate}
Early termination is one of the essential components in the training procedure \cite{peng2018deepmimic}, which eliminates the local minimum or corner case with unnatural performance.
In our manipulation task, the early termination would be triggered when:
\begin{itemize}
  \item The robot gets in self-collision.
  \item The ball contacts with other links except for the end-effector on feet.
  \item The ball's position is out of a feasible region.
  \item The period in which the ball is in a no-contact state exceeds a threshold.
\end{itemize}
The threshold region is set as a horizontal plane being $\pm{1.5}$ radius of the ball and vertical axis being $\pm{1}$ radius.
The limit on ball-no-contact time is to avoid solutions in which the robot throws the rotating ball into the air.

\subsection{Domain Randomization}
\label{sec:method-rand}

\begin{figure}[t]
  \begin{center}
    \centering
    \includegraphics[clip,  bb= 0 0 960 390,  width=1.0\columnwidth]{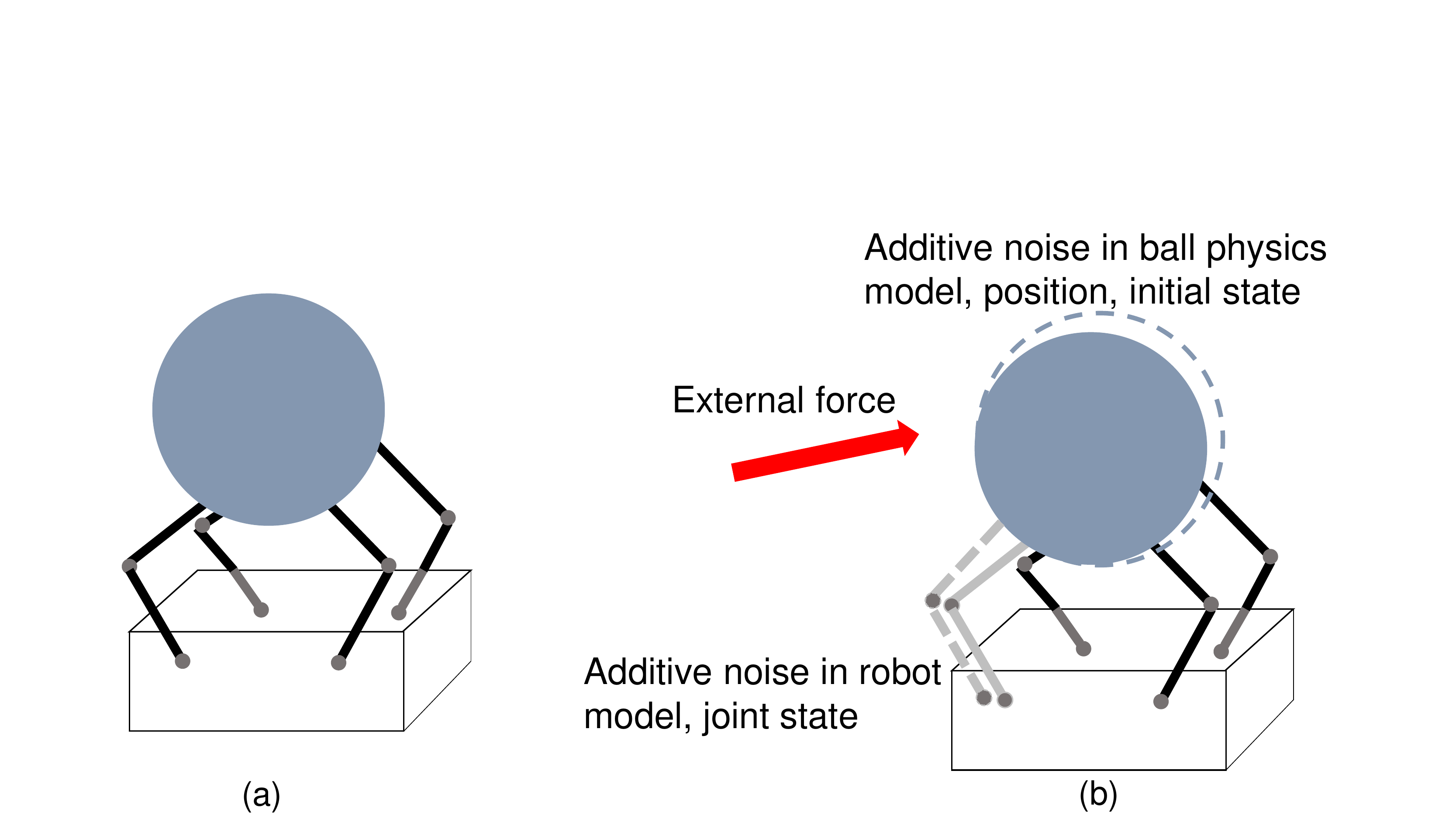}
    \caption{Visualization of domain randomization and external disturbance force during training.}
    \label{figure:domain-rand}
  \end{center}
\end{figure}

Domain randomization is a simple technique to improve a policy's robustness against modeling errors and sensor measurement noise as \figref{figure:domain-rand}.
In our circus task, the randomization takes place in the followings:
\begin{itemize}
  \item Leg configuration. The shank positions and lengths are perturbed with additive noise $\sim$ $\scalebox{0.9}{\unit[$\mathcal{N}(0.03, 0)]{m}$}$.
  \item Joint state observation. Additive noise $\sim$ $\scalebox{0.9}{$\unit[\mathcal{N}(0.05, 0)]{rad}$}$ to joint positions and  $\sim$ $\scalebox{0.9}{$\unit[\mathcal{N}(0.3, 0)]{rad/s}$}$ to velocities. 
  \item Ball physical model, including mass with a random $\scalebox{0.9}{\unit[$\nu(\pm5)]{\%}$}$ of its original weight and radius with $\scalebox{0.9}{\unit[$\nu(\pm10)]{\%}$}$
  \item Ball contact model. The friction and restitution coefficients are sampled from uniform distributions $U(0.5, 1.1)$ and $U(0.9, 1.0)$
  \item Ball state observation. Additive noise $\sim$ $\scalebox{0.9}{$\unit[\mathcal{N}(0.04, 0)]{m}$}$ to position and $\sim$ $\scalebox{0.9}{$\unit[\mathcal{N}(0.03, 0)]{rad }$}$ to orientation in each axis
  \item Initialization state, including initial robot position and configuration, ball position and orientation
\end{itemize}
where the policy is encouraged to learn strategies under these varying dynamics and initialization to better deal with real world conditions.

\subsection{External Disturbance}
\label{sec:method-disturb}
Injecting random disturbance force has shown to be effective in achieving sim-to-real transfer \cite{lee2020sci, valassakis2020force_injection}.
During training, the ball is applied with \unit[50]{N} external force from a random direction.
The disturbance force lasts for \unit[0.4]{s} and appears at the random time point with \unit[20]{\%} probability.


\subsection{Policy Training}
\label{sec:method-alg}
\begin{figure}[t]
  \begin{center}
    \includegraphics[clip,  bb= 10 0 920 355,  width=1.0\columnwidth]{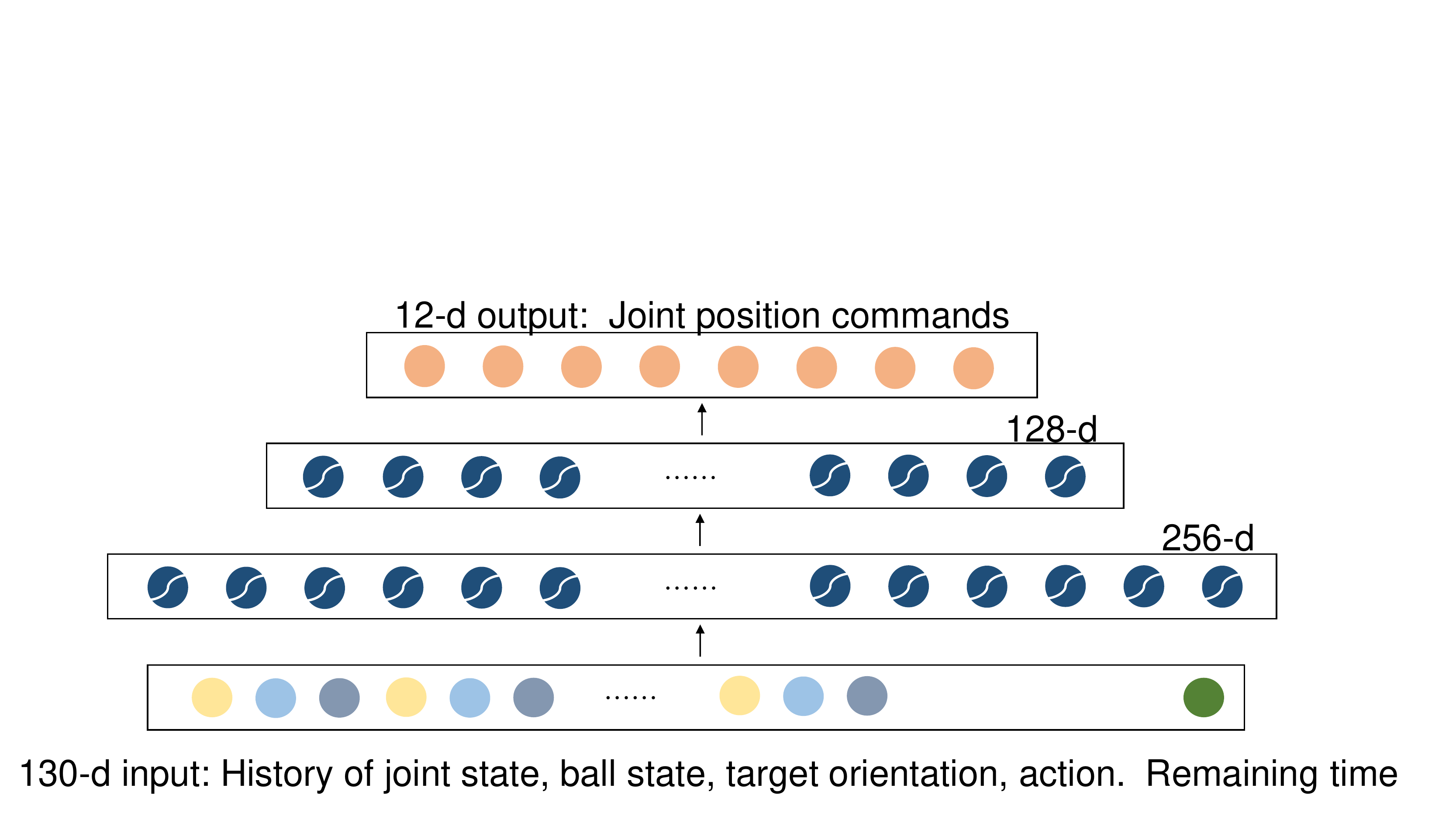}
    \caption{Network architecture used for policy is a two-layers Multi-Layer Perceptron (MLP) with tanh activation function. The output is directly send to robot as joints position commands. The value network is similar architecture but with $1$d output value.}
    \label{figure:network}
  \end{center}
\end{figure}

The policy and value network are multi-layer perceptron (MLP) with $2$ hidden layers with 256 and 128 units for each and tanh activation function as \figref{figure:network}.

The Proximal Policy Optimization (PPO) \cite{schulman2017ppo} algorithm is used for training. The hyperparameters are: 
discount factor $\scalebox{1.0}{$\gamma = 0.998$}$, clipping range $\scalebox{1.0}{$\epsilon = 0.2$}$, learning rate $\scalebox{1.0}{$lr = 0.001$}$.
The parameterized policy $\pi_{\theta}(a_t | o_t)$ is used to maximize the expected reward return:
\begin{equation}
  \label{eq:reward_return}
  \scalebox{0.9}{$
    \theta^{*} = \argmax_{\theta} \mathbb{E}_{\pi_{\theta}}\left[
      \displaystyle
      \sum_{t=0}^{\infty} \gamma^{t}(r_{t}^{q} + r_{t}^{v} + r_{t}^{\tau} + r_{t}^{slip} + r_{t}^{collide})
      \right]
  $}
\end{equation}

\subsection{Curriculum Learning}
\label{sec:method-curriculum}

Curriculum learning \cite{bengio2009curriculum} is an effective strategy to solve challenging tasks,
in which the task difficulties gradually increase during the learning process.
Such a technique has proven to be useful in locomotion tasks where the terrain's complexity is gradually increased in the course of training, resulting in a higher survival time in training rollouts \cite{xie2020allsteps,lee2020sci}.
In this work, the curricula focus on the domain randomization ranges, target rotation speed, and discrete period for updating target orientation.
The robot is initially trained with the low-speed manipulation commands and small additive noise in domain randomization.
The target speed and noise would increase monotonically, while the updating period decreases.
The initial target speed is $\unit[0]{^{\circ}/s}$ and the final speed is $\unit[15]{^{\circ}/s}$.
The updating period is set as discrete value from $1$, $0.5$ to $0.33$ s,
in which the target orientation is updated at $1$, $2$ and \unit[3]{Hz}.
To maximize total reward, the policy would generate the higher-frequency gait when the updating period decreases.
The modeling and additive noise in \secref{method-rand} is discounted with a curriculum factor whose value would increase to $1$ in the end.
\section{Experiment}
\label{sec:result}

\begin{figure*}[t]
  \begin{center}
    \includegraphics[clip,  bb= 0 0 960 540, width=1.0\columnwidth]{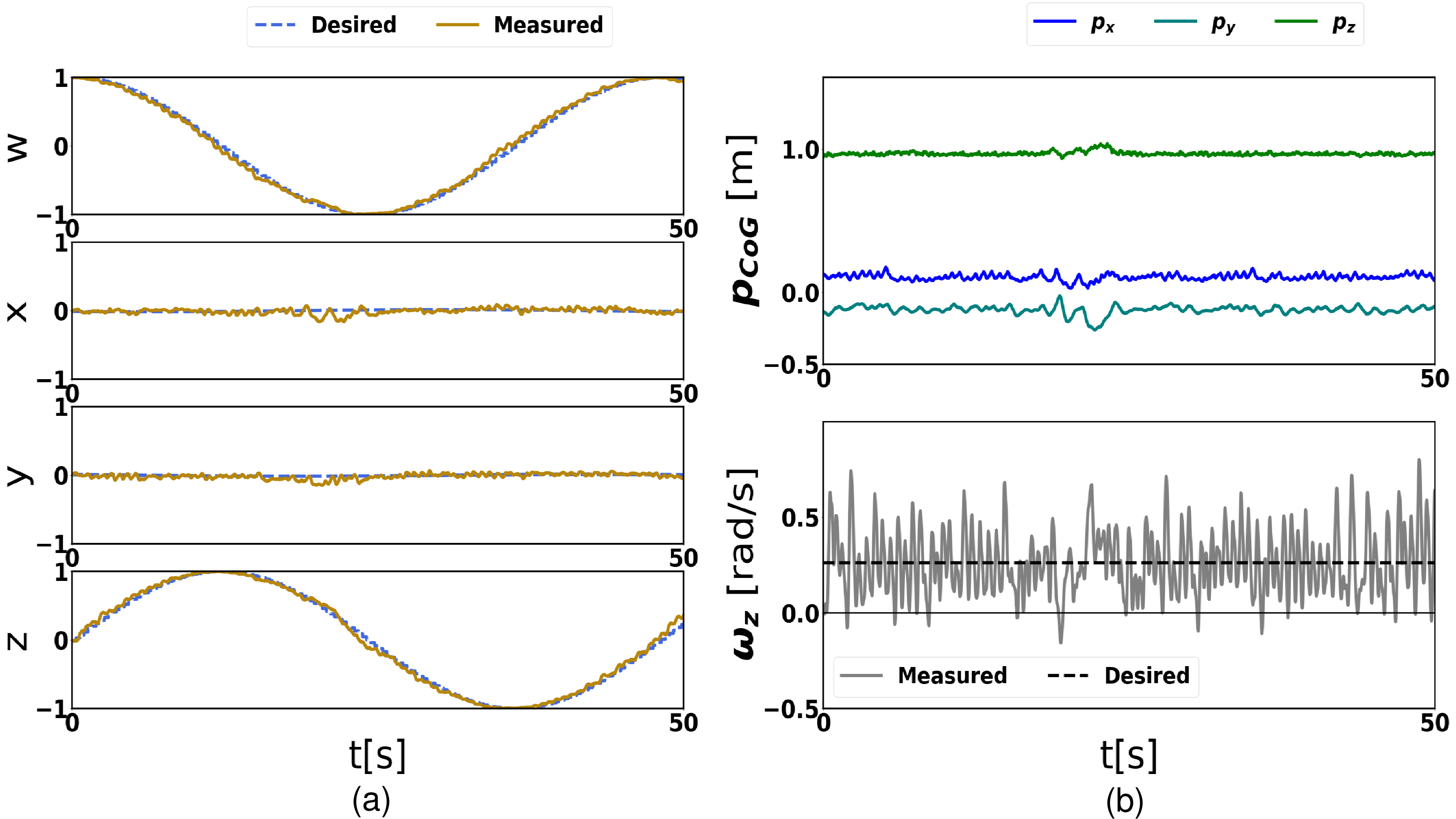}
    \includegraphics[clip,  bb= 0 0 960 540,  width=1.0\columnwidth]{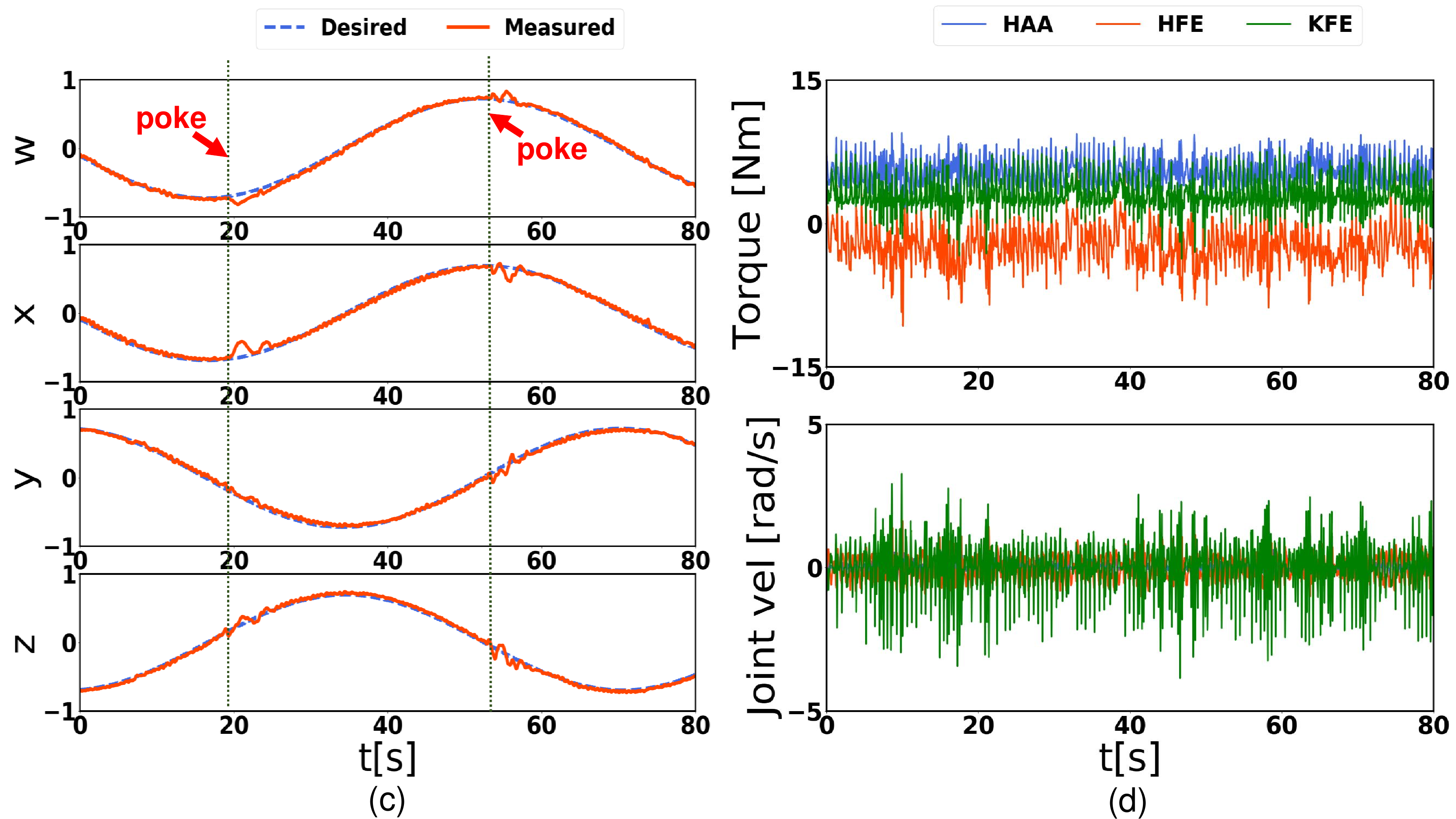}

    \caption{Tracking results of different direction and different speeds on the real robot, in which (a), (b) show a yaw-rotation experiment with $\unit[15]{^{\circ}/s}$ speed, and (c), (d) show a pitch-rotation experiment with $\unit[10]{^{\circ}/s}$ speed under the external disturbance. Among them, (a) is the quaternion tracking results; (b) is the position and angular velocity of the ball, which are measured by the motion capture system; (c) is the quaternion tracking results under external pokes, in which the second poke is snapshotted as \figref{figure:result-pitch-poke};
    (d) is the joint torque and velocity during the manipulation. }
    \label{figure:result-quat-pitch-yaw}
  \end{center}
\end{figure*}

The policy is trained in simulation and achieves zero-shot transfer to the real robot.
We demonstrate our framework on ANYmal-C\footnote{https://www.anybotics.com} with a total mass of about \unit[$40$]{kg} and $12$ SEAs with a maximum torque of \unit[$80$]{Nm}.
The ball in the experiment is a commercial yoga ball with \unit[3]{kg} mass and \unit[0.8]{m} diameter.
A VICON system\footnote{https://www.vicon.com} measures
Center-of-gravity (CoG) position and orientation of the ball.

We have identified the following mismatches between simulation and the real robot experiments: 
the joints position tracking errors, the softness and deformation in yoga ball, 
the unexpected slip between ball surface and feet, 
the non-standard ellipse shape after inflation and CoG position error.
However, our learned control policy has proven to be robust to these uncertainties.

To avoid the robot's feet colliding with the motion capture markers,
we separately demonstrate the rotation in roll, pitch, and yaw direction with a maximum rotational speed of $\unit[15]{^{\circ}/s}$.
In the real robot, we conducted manipulation experiments for over $\unit[2]{mins}$, during which the controller could recover robustly under external disturbance such as pocking the ball.

\subsection{Simulation}
\label{sec:result-sim}
The policy is trained in the Raisim simulator \cite{hwangbo2018raisim}, which is used to simulate the rigid-body contact dynamics.
During training, an actuator network \cite{hwangbo2019scirob} is used to reduce the modeling mismatch between simulation and real robots due to unmodeled actuator dynamics.

The quality of the trained policy is strongly related to both target speeds and updating periods of the target orientation.
Larger target speed leads to more aggressive motion,
while a shorter period leads to more frequent contact changes.
The policy is trained with different updating periods, as described in~\secref{method-curriculum}.
The simulation results show that the policy with an extended period generates smoother motion with less contact switch but easier to fail when the target speed and domain randomization noises are increased.
In contrast, the policy with a short period leads to more contact switch, but enables larger target speed with more robustness.
Due to the joints' physical limitations in velocity and acceleration, the updating period could not be infinitely small.
Analogous to the gait scheduling settings in locomotion, the final period is selected to be \unit[0.33]{s} by updating the target orientation at \unit[3]{Hz}.

We also notice that policy becomes more conservative with increasing external disturbance and growing domain randomization.
One example is foot clearance decreases during training to quickly respond to the unexpected disturbance.

\subsection{Real Robot}
\label{sec:result-real}
We run the same policy in the real robot without any modification.
The policy runs \unit[100]{Hz} on the real robot and sends joints position command to the robot's joint position PD controllers.

In the real experiment, we evaluate the roll, pitch, and yaw rotation separately.
For each direction, the target rotation speed is set as $\unit[10]{^{\circ}/s}$ and $\unit[15]{^{\circ}/s}$ to demonstrate the dynamic performance.
To verify the robustness of the policy, we randomly poke the ball and robot to simulate external disturbances during the experiments.

\begin{figure*}[t]
  \begin{center}
    \includegraphics[clip,  bb= 0 210 960 410,  width=1.8\columnwidth]{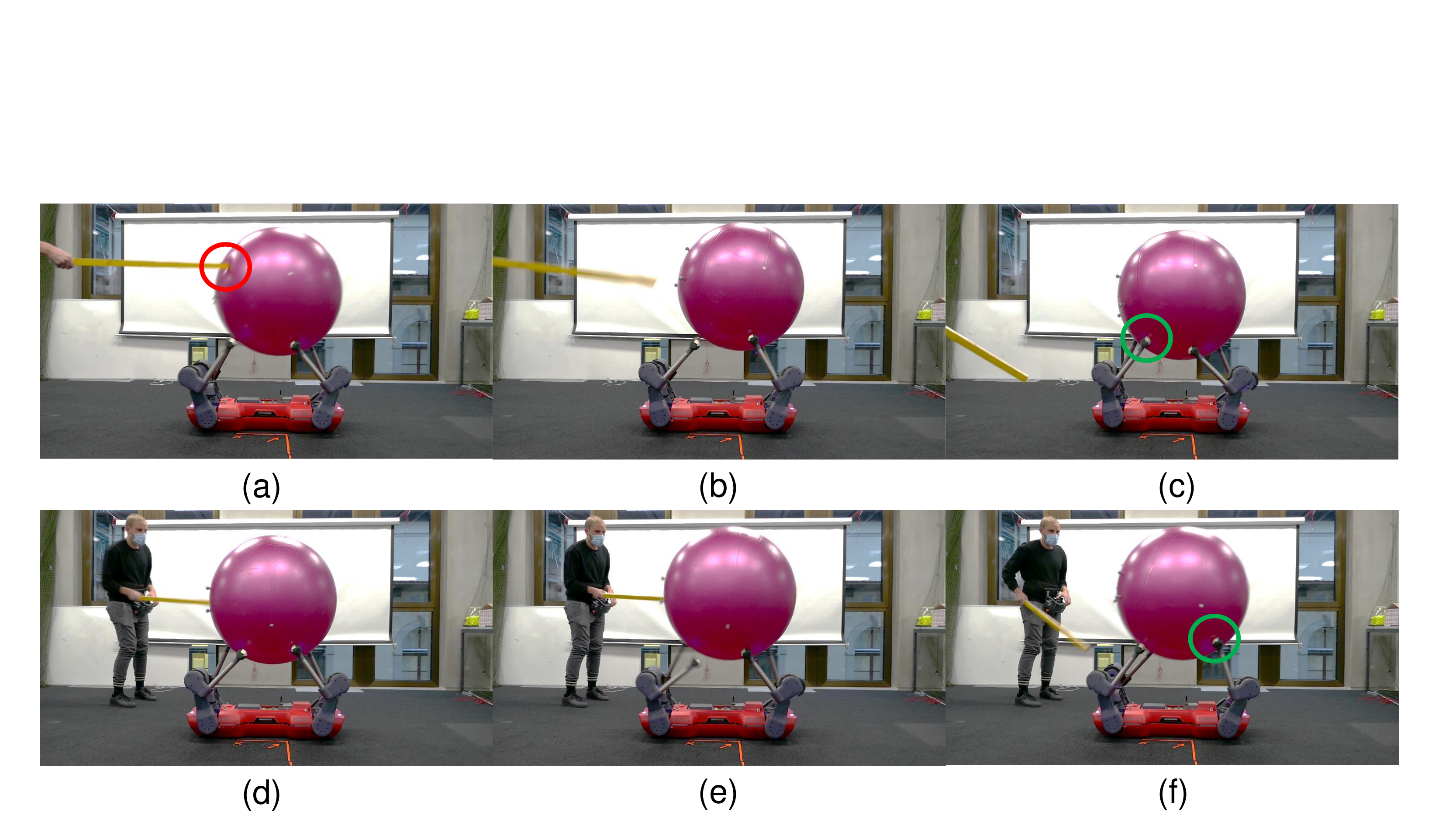}
    \caption{Recovery from the ball being poked,  while the robot is commanded to manipulate in the roll direction with $\unit[15]{^{\circ}/s}$. 
    (a) shows the ball being poked, in which the deformation could be noticed obviously; (b) shows the ball's state after poking; (c) shows the recovery motion generated by the policy. }
    \label{figure:result-roll-poke}
  \end{center}
\end{figure*}

\begin{figure*}[t]
  \begin{center}
    \includegraphics[clip,  bb= 0 0 960 410,  width=1.8\columnwidth]{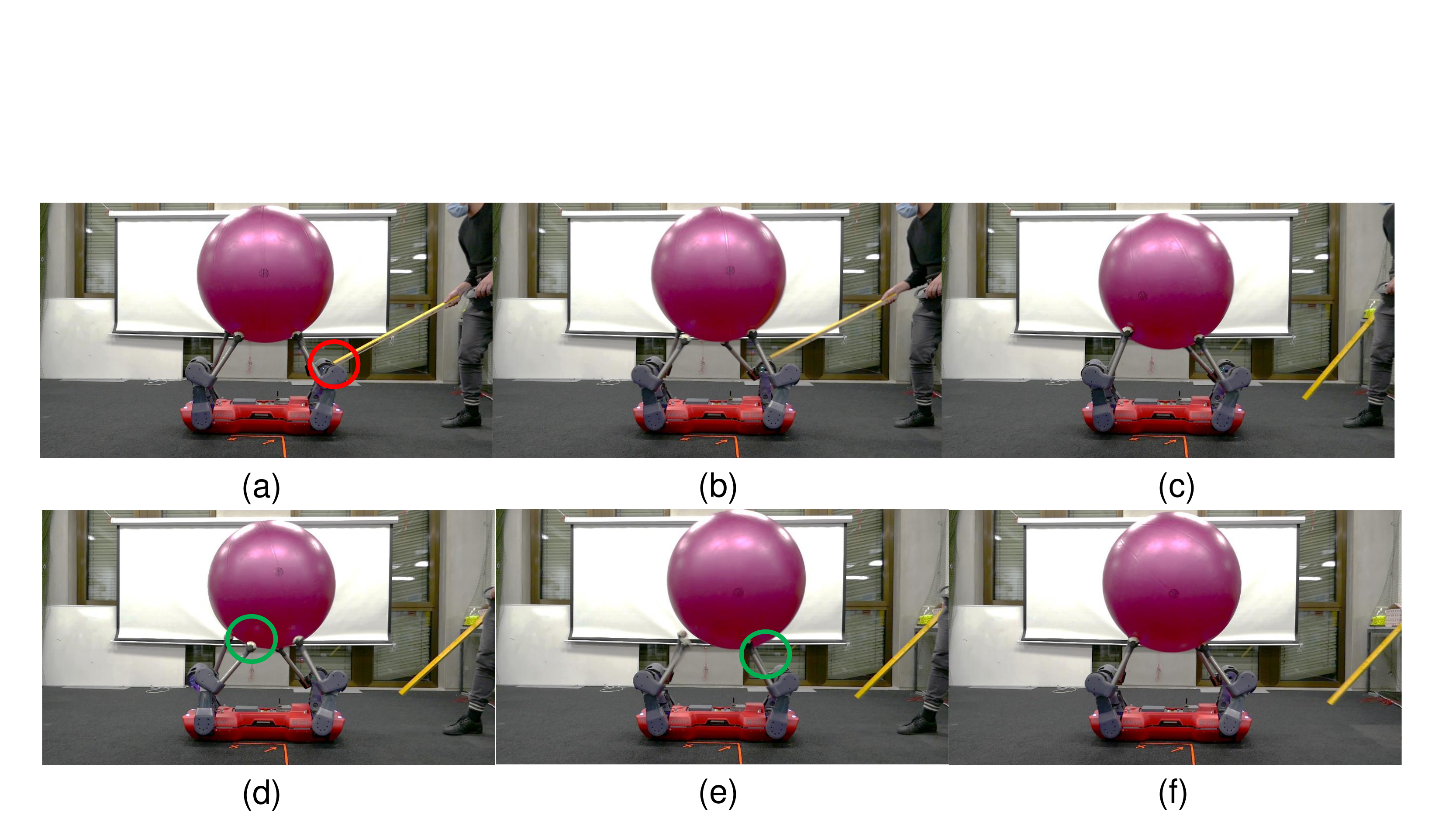}
    \caption{Recovery from the robot leg being poked,  while the robot is commanded to manipulate in the pitch direction with $\unit[10]{^{\circ}/s}$.
    (a), (b) show the leg is poked with an apparent displacement in the end-effector; (c) shows influences of the contact's displacement because of the poking, in which the ball's orientation varies abnormally; (d), (e) show the recovery motion generated by the policy; (f) shows the final state after the recovery. }
    \label{figure:result-pitch-poke}
  \end{center}
\end{figure*}

The tracking performance is shown in \figref{figure:result-quat-pitch-yaw}, in which the results of accurate tracking and recovery from the external force are presented.
We randomly poke the ball and robot limbs from different directions, as the snapshots in \figref{figure:result-roll-poke} and \figref{figure:result-pitch-poke}.
Our robot quickly recovers and continues the task, though we do not have a high-level task planner, and our non-prehensile style is less stable.
Moreover, we notice in \figref{figure:result-quat-pitch-yaw}(b) that the angular velocity fluctuates from the target value, even if the orientation is tracked pretty well.
As \figref{figure:result-quat-pitch-yaw}(d), the peak and averages of joints' torque and velocities during manipulation are about $\frac{1}{4}$ of locomotion tasks using a similar ANYmal robot \cite{hwangbo2019scirob}.

As for larger rotation speeds, we realize that
when the target velocity is over $\unit[20]{^{\circ}/s}$,
there is more severe slippage between feet and ball because of the under-actuated legs design and limited contact area, which drives the system more likely to fail.
When speeds become more enormous, the joints motion becomes more aggressive with higher acceleration, leading to a large force on the robot against the ground, which is more likely to damage the robot base. 

\section{Discussion}
\label{sec:discussion}

\subsection{Locomotion and Manipulation}
\label{sec:dis-locomanip}
Multi-legged locomotion shows the duality with multi-fingered manipulation \cite{raibert1986legged, lynch1996nonprehensile, johnson2013selfmanip}.
This section compares the similarity and difference between locomotion and our manipulation task under the reinforcement learning settings.

\paragraph{Speed Reward}
In locomotion, the reward mostly directly depends on velocity \cite{hwangbo2019scirob, lee2019robust, yang2020data, da2020learning}.
In contrast for our manipulation task, we discretize the time at a fixed frequency and update the target quaternion value based on the velocity command.
Compared to direct velocity-based reward, we found that our reward design converges better during training.
We speculate that the main reason is that the terrain is more complicated in our manipulation case than during flat ground locomotion, making it more difficult to track the constant speed. A similar observation was made for rough-terrain locomotion~\cite{lee2020sci}. 
As it is shown in the experimental results \figref{figure:result-quat-pitch-yaw}(b), the average velocity is pretty accurate, while the speed can be quite noisy.
We also study humans conducting a similar task with a soccer ball and using their fingers. While establishing a direct parallel is impossible, we observe resembling undulant motions during the task execution.  

\paragraph{Foot Clearance}
Foot clearance denotes the distance between legs end-effector and the environment.
It is a popular reward term in shaping gait behavior during locomotion to avoid foot scuffing~\cite{hwangbo2019scirob, lee2019robust, lee2020sci}.
The reward is often calculated by variations to a heuristic clearance value.
In our manipulation task, foot clearance term is removed because it is not intuitive to decide on a heuristic value.

\paragraph{Gait Pattern}
In a quadrupedal locomotion task, the gait pattern is often preset with specific contact sequences provided as input to the learned policy \cite{peng2020dog, tan2018sim, yang2020data, ha2020learning}.
In our work, we do not define any gait pattern. In contrast to legged locomotion, where the heuristics can be directly inspired by well-studied animal gaits, we cannot rely on comparable gait information for in-limb manipulation.

\paragraph{Model-based and Model-free}
In quadrupedal locomotion tasks, model-based RL combines learned gait with a model-based whole-body controller \cite{da2020learning}.
However, due to the whole-body controller's sensitivity to model-mismatches, it was not suitable for our task as the ball deformation and slippage impede any accurate modeling.

\subsection{In-Limb vs Dexterous Hand Manipulation}
\label{sec:dis-quad-hand}
Our task is similar to in-hand manipulation with a dexterous robot hand.
However, the fingers in the robot hand are usually driven by compact actuators with limited torque output and joint velocity.
On the contrary, the actuators in our quadrupedal robot are much more powerful with respect to the manipulandum's weight.
The maximum torque in each SEA motor is $\unit[80]{Nm}$, while the torque usage during manipulation is less than $\unit[5]{Nm}$ in the most time as \figref{figure:result-quat-pitch-yaw}(d).
This difference in relative actuator power influences the reward design. While previous RL-based in-hand dexterous manipulation researches could design the reward with only target orientation term \cite{vikash2019dexterous, openai2020inhand, openai2019rubik}, our setup requires accounting for limbs actuation. 
We have realised that without penalizing the control effort, the trained policy would generate an overly aggressive motion with large joint velocity and acceleration without punishment in the reward design. 

\section{Conclusion}
\label{sec:conclusion}
This paper has proposed a novel and robust approach based on deep reinforcement learning for the quadrupedal robot to achieve a first step towards dexterous full-limbs manipulation.
The policy is trained in the simulation with model-free RL, and achieve zero-shot deployment on the real robot.
This is, to our best knowledge, the first work to achieve dexterous dynamic manipulation on the real quadrupedal robot.

The proposed controller exhibits dynamic manipulation performance and achieves a maximum $15 deg/s$ ball rotation speed on hardware experiments. 
The controller's robustness is verified on hardware by showing first, a continuous manipulation for more than $2$ minutes, and second, a robust recovery under external disturbances during manipulation.

In the end, we revisit the classical argument of the duality between manipulation and locomotion.
By comparing the similarities and differences in reward design under RL settings, for the challenging terrain, we hope to inspire a more closed connection on both sides.

\section*{Acknowledgments}
This research was supported by the Swiss National Science Foundation through the National Center of Competence in Research (NCCR) Robotics.
We appreciate the insightful discussion with Vassilios Tsounis, Bowen Yang, Jan Carius, Lorenz Wellhausen from ETH Z\"urich, Prof. Davide Scaramuzza from the University of Z\"urich, Jemin Hwangbo from KAIST, Hironori Yoshida from the University of Tokyo, Yifan Hou from CMU, Prof. Weiwei Wan from Osaka University.

\bibliography{0-main}

\begin{thebibliography}{10}
\providecommand{\url}[1]{#1}
\csname url@samestyle\endcsname
\providecommand{\newblock}{\relax}
\providecommand{\bibinfo}[2]{#2}
\providecommand{\BIBentrySTDinterwordspacing}{\spaceskip=0pt\relax}
\providecommand{\BIBentryALTinterwordstretchfactor}{4}
\providecommand{\BIBentryALTinterwordspacing}{\spaceskip=\fontdimen2\font plus
\BIBentryALTinterwordstretchfactor\fontdimen3\font minus
  \fontdimen4\font\relax}
\providecommand{\BIBforeignlanguage}[2]{{%
\expandafter\ifx\csname l@#1\endcsname\relax
\typeout{** WARNING: IEEEtran.bst: No hyphenation pattern has been}%
\typeout{** loaded for the language `#1'. Using the pattern for}%
\typeout{** the default language instead.}%
\else
\language=\csname l@#1\endcsname
\fi
#2}}
\providecommand{\BIBdecl}{\relax}
\BIBdecl

\bibitem{sangbae2019mini}
B.~Katz, J.~Di~Carlo, and S.~Kim, ``Mini cheetah: A platform for pushing the
  limits of dynamic quadruped control,'' in \emph{2019 International Conference
  on Robotics and Automation (ICRA)}.\hskip 1em plus 0.5em minus 0.4em\relax
  IEEE, 2019, pp. 6295--6301.

\bibitem{bdspot}
``Boston dynamics spot,'' https://www.bostondynamics.com/spot, accessed:
  2020-10-10.

\bibitem{hwangbo2019scirob}
J.~Hwangbo, J.~Lee, A.~Dosovitskiy, D.~Bellicoso, V.~Tsounis, V.~Koltun, and
  M.~Hutter, ``Learning agile and dynamic motor skills for legged robots,''
  \emph{Science Robotics}, vol.~4, no.~26, 2019.

\bibitem{lee2019robust}
J.~Lee, J.~Hwangbo, and M.~Hutter, ``Robust recovery controller for a
  quadrupedal robot using deep reinforcement learning,'' \emph{arXiv preprint
  arXiv:1901.07517}, 2019.

\bibitem{tsounis2020deepgait}
V.~Tsounis, M.~Alge, J.~Lee, F.~Farshidian, and M.~Hutter, ``Deepgait: Planning
  and control of quadrupedal gaits using deep reinforcement learning,''
  \emph{IEEE Robotics and Automation Letters}, vol.~5, no.~2, pp. 3699--3706,
  2020.

\bibitem{da2020learning}
X.~Da, Z.~Xie, D.~Hoeller, B.~Boots, A.~Anandkumar, Y.~Zhu, B.~Babich, and
  A.~Garg, ``Learning a contact-adaptive controller for robust, efficient
  legged locomotion,'' \emph{arXiv preprint arXiv:2009.10019}, 2020.

\bibitem{lee2020sci}
\BIBentryALTinterwordspacing
J.~Lee, J.~Hwangbo, L.~Wellhausen, V.~Koltun, and M.~Hutter, ``Learning
  quadrupedal locomotion over challenging terrain,'' \emph{Science Robotics},
  vol.~5, no.~47, 2020. [Online]. Available:
  \url{https://robotics.sciencemag.org/content/5/47/eabc5986}
\BIBentrySTDinterwordspacing

\bibitem{Bellicoso2018jfr}
\BIBentryALTinterwordspacing
C.~D. Bellicoso, M.~Bjelonic, L.~Wellhausen, K.~Holtmann, F.~G{\"{u}}nther,
  M.~Tranzatto, P.~Fankhauser, and M.~Hutter, ``{Advances in real-world
  applications for legged robots},'' \emph{Journal of Field Robotics}, vol.~35,
  no.~8, pp. 1311--1326, dec 2018. [Online]. Available:
  \url{http://doi.wiley.com/10.1002/rob.21839}
\BIBentrySTDinterwordspacing

\bibitem{jpl2020subt}
A.~Bouman, M.~F. Ginting, N.~Alatur, M.~Palieri, D.~D. Fan, T.~Touma,
  T.~Pailevanian, S.-K. Kim, K.~Otsu, J.~Burdick \emph{et~al.}, ``Autonomous
  spot: Long-range autonomous exploration of extreme environments with legged
  locomotion,'' \emph{arXiv preprint arXiv:2010.09259}, 2020.

\bibitem{iit2016quad_plus_arm}
B.~U. Rehman, M.~Focchi, J.~Lee, H.~Dallali, D.~G. Caldwell, and C.~Semini,
  ``Towards a multi-legged mobile manipulator,'' in \emph{2016 IEEE
  International Conference on Robotics and Automation (ICRA)}.\hskip 1em plus
  0.5em minus 0.4em\relax IEEE, 2016, pp. 3618--3624.

\bibitem{rsl2019alma}
C.~D. Bellicoso, K.~Kr{\"a}mer, M.~St{\"a}uble, D.~Sako, F.~Jenelten,
  M.~Bjelonic, and M.~Hutter, ``Alma-articulated locomotion and manipulation
  for a torque-controllable robot,'' in \emph{2019 International Conference on
  Robotics and Automation (ICRA)}.\hskip 1em plus 0.5em minus 0.4em\relax IEEE,
  2019, pp. 8477--8483.

\bibitem{sethu2020anymal_arm}
H.~Ferrolho, W.~Merkt, V.~Ivan, W.~Wolfslag, and S.~Vijayakumar, ``Optimizing
  dynamic trajectories for robustness to disturbances using polytopic
  projections,'' \emph{arXiv preprint arXiv:2003.00609}, 2020.

\bibitem{mori_2002_whole_quad_manip}
T.~Omata, K.~Tsukagoshi, and O.~Mori, ``Whole quadruped manipulation,'' in
  \emph{Proceedings 2002 IEEE International Conference on Robotics and
  Automation (Cat. No. 02CH37292)}, vol.~2.\hskip 1em plus 0.5em minus
  0.4em\relax IEEE, 2002, pp. 2028--2033.

\bibitem{hong2020transport}
J.~{Hooks}, M.~S. {Ahn}, J.~{Yu}, X.~{Zhang}, T.~{Zhu}, H.~{Chae}, and
  D.~{Hong}, ``Alphred: A multi-modal operations quadruped robot for package
  delivery applications,'' \emph{IEEE Robotics and Automation Letters}, vol.~5,
  no.~4, pp. 5409--5416, 2020.

\bibitem{noriho_1995_hexapod_manip}
N.~Koyachi, T.~Arai, H.~Adachi, K.-i. Asami, and Y.~Itoh, ``Hexapod with
  integrated limb mechanism of leg and arm,'' in \emph{Proceedings of 1995 IEEE
  International Conference on Robotics and Automation}, vol.~2.\hskip 1em plus
  0.5em minus 0.4em\relax IEEE, 1995, pp. 1952--1957.

\bibitem{noriho_2000_hexapod_rise}
Y.~Takahashi, T.~Arai, Y.~Mae, K.~Inoue, and N.~Koyachi, ``Development of
  multi-limb robot with omnidirectional manipulability and mobility,'' in
  \emph{Proceedings. 2000 IEEE/RSJ International Conference on Intelligent
  Robots and Systems (IROS 2000)(Cat. No. 00CH37113)}, vol.~3.\hskip 1em plus
  0.5em minus 0.4em\relax IEEE, 2000, pp. 2012--2017.

\bibitem{adachi_2002_hexapod_manip}
N.~Koyachi, H.~Adachi, M.~Izumi, and T.~Hirose, ``Control of walk and
  manipulation by a hexapod with integrated limb mechanism: Melmantis-1,'' in
  \emph{Proceedings 2002 IEEE International Conference on Robotics and
  Automation (Cat. No. 02CH37292)}, vol.~4.\hskip 1em plus 0.5em minus
  0.4em\relax IEEE, 2002, pp. 3553--3558.

\bibitem{inoue2010pushing}
K.~Inoue, K.~Ooe, and S.~Lee, ``Pushing methods for working six-legged robots
  capable of locomotion and manipulation in three modes,'' in \emph{2010 IEEE
  International Conference on Robotics and Automation}.\hskip 1em plus 0.5em
  minus 0.4em\relax IEEE, 2010, pp. 4742--4748.

\bibitem{Vijayakumar2020anymal_manip}
W.~Wolfslag, C.~McGreavy, G.~Xin, C.~Tiseo, S.~Vijayakumar, and Z.~Li,
  ``Optimisation of body-ground contact for augmenting whole-body
  loco-manipulation of quadruped robots,'' \emph{arXiv preprint
  arXiv:2002.10552}, 2020.

\bibitem{ma2020tro}
G.~{Zhang}, S.~{Ma}, Y.~{Shen}, and Y.~{Li}, ``A motion planning approach for
  nonprehensile manipulation and locomotion tasks of a legged robot,''
  \emph{IEEE Transactions on Robotics}, vol.~36, no.~3, pp. 855--874, 2020.

\bibitem{ucb2020quad_ball}
C.~Yang, B.~Zhang, J.~Zeng, A.~Agrawal, and K.~Sreenath, ``Dynamic legged
  manipulation of a ball through multi-contact optimization,'' \emph{arXiv
  preprint arXiv:2008.00191}, 2020.

\bibitem{raibert1986legged}
M.~H. Raibert, \emph{Legged robots that balance}.\hskip 1em plus 0.5em minus
  0.4em\relax MIT press, 1986.

\bibitem{lynch1996nonprehensile}
K.~M. Lynch, \emph{Nonprehensile robotic manipulation: Controllability and
  planning}.\hskip 1em plus 0.5em minus 0.4em\relax Citeseer, 1996.

\bibitem{johnson2013selfmanip}
A.~M. Johnson and D.~E. Koditschek, ``Legged self-manipulation,'' \emph{IEEE
  Access}, vol.~1, pp. 310--334, 2013.

\bibitem{aiyama_pivot}
Y.~Aiyama, M.~Inaba, and H.~Inoue, ``Pivoting: A new method of graspless
  manipulation of object by robot fingers,'' in \emph{Proceedings of 1993
  IEEE/RSJ International Conference on Intelligent Robots and Systems
  (IROS'93)}, vol.~1.\hskip 1em plus 0.5em minus 0.4em\relax IEEE, 1993, pp.
  136--143.

\bibitem{okamura2000dexterous_survey}
A.~M. Okamura, N.~Smaby, and M.~R. Cutkosky, ``An overview of dexterous
  manipulation,'' in \emph{Proceedings 2000 ICRA. Millennium Conference. IEEE
  International Conference on Robotics and Automation. Symposia Proceedings
  (Cat. No. 00CH37065)}, vol.~1.\hskip 1em plus 0.5em minus 0.4em\relax IEEE,
  2000, pp. 255--262.

\bibitem{bicchi2000survey}
A.~Bicchi and V.~Kumar, ``Robotic grasping and contact: A review,'' in
  \emph{Proceedings 2000 ICRA. Millennium Conference. IEEE International
  Conference on Robotics and Automation. Symposia Proceedings (Cat. No.
  00CH37065)}, vol.~1.\hskip 1em plus 0.5em minus 0.4em\relax IEEE, 2000, pp.
  348--353.

\bibitem{mordatch2012dexterous_sim}
I.~Mordatch, Z.~Popovi{\'c}, and E.~Todorov, ``Contact-invariant optimization
  for hand manipulation,'' in \emph{Proceedings of the ACM
  SIGGRAPH/Eurographics symposium on computer animation}, 2012, pp. 137--144.

\bibitem{dollar2012dexterous}
I.~M. Bullock, R.~R. Ma, and A.~M. Dollar, ``A hand-centric classification of
  human and robot dexterous manipulation,'' \emph{IEEE transactions on
  Haptics}, vol.~6, no.~2, pp. 129--144, 2012.

\bibitem{karen2014dexterous_sim}
Y.~Bai and C.~K. Liu, ``Dexterous manipulation using both palm and fingers,''
  in \emph{2014 IEEE International Conference on Robotics and Automation
  (ICRA)}.\hskip 1em plus 0.5em minus 0.4em\relax IEEE, 2014, pp. 1560--1565.

\bibitem{kumar2016optimal_dexterous}
V.~Kumar, E.~Todorov, and S.~Levine, ``Optimal control with learned local
  models: Application to dexterous manipulation,'' in \emph{2016 IEEE
  International Conference on Robotics and Automation (ICRA)}.\hskip 1em plus
  0.5em minus 0.4em\relax IEEE, 2016, pp. 378--383.

\bibitem{van2015learning}
H.~Van~Hoof, T.~Hermans, G.~Neumann, and J.~Peters, ``Learning robot in-hand
  manipulation with tactile features,'' in \emph{2015 IEEE-RAS 15th
  International Conference on Humanoid Robots (Humanoids)}.\hskip 1em plus
  0.5em minus 0.4em\relax IEEE, 2015, pp. 121--127.

\bibitem{vikash2017learning}
A.~Rajeswaran, V.~Kumar, A.~Gupta, G.~Vezzani, J.~Schulman, E.~Todorov, and
  S.~Levine, ``Learning complex dexterous manipulation with deep reinforcement
  learning and demonstrations,'' \emph{arXiv preprint arXiv:1709.10087}, 2017.

\bibitem{barth2018dexterous_sim}
G.~Barth-Maron, M.~W. Hoffman, D.~Budden, W.~Dabney, D.~Horgan, D.~Tb,
  A.~Muldal, N.~Heess, and T.~Lillicrap, ``Distributed distributional
  deterministic policy gradients,'' \emph{arXiv preprint arXiv:1804.08617},
  2018.

\bibitem{plappert2018dexterous_sim}
M.~Plappert, M.~Andrychowicz, A.~Ray, B.~McGrew, B.~Baker, G.~Powell,
  J.~Schneider, J.~Tobin, M.~Chociej, P.~Welinder \emph{et~al.}, ``Multi-goal
  reinforcement learning: Challenging robotics environments and request for
  research,'' \emph{arXiv preprint arXiv:1802.09464}, 2018.

\bibitem{vikash2019dexterous}
H.~Zhu, A.~Gupta, A.~Rajeswaran, S.~Levine, and V.~Kumar, ``Dexterous
  manipulation with deep reinforcement learning: Efficient, general, and
  low-cost,'' in \emph{2019 International Conference on Robotics and Automation
  (ICRA)}.\hskip 1em plus 0.5em minus 0.4em\relax IEEE, 2019, pp. 3651--3657.

\bibitem{openai2019rubik}
I.~Akkaya, M.~Andrychowicz, M.~Chociej, M.~Litwin, B.~McGrew, A.~Petron,
  A.~Paino, M.~Plappert, G.~Powell, R.~Ribas \emph{et~al.}, ``Solving rubik's
  cube with a robot hand,'' \emph{arXiv preprint arXiv:1910.07113}, 2019.

\bibitem{openai2020inhand}
O.~M. Andrychowicz, B.~Baker, M.~Chociej, R.~Jozefowicz, B.~McGrew,
  J.~Pachocki, A.~Petron, M.~Plappert, G.~Powell, A.~Ray \emph{et~al.},
  ``Learning dexterous in-hand manipulation,'' \emph{The International Journal
  of Robotics Research}, vol.~39, no.~1, pp. 3--20, 2020.

\bibitem{vikash2020inhand_ball}
A.~Nagabandi, K.~Konolige, S.~Levine, and V.~Kumar, ``Deep dynamics models for
  learning dexterous manipulation,'' in \emph{Conference on Robot Learning},
  2020, pp. 1101--1112.

\bibitem{tan2018sim}
J.~Tan, T.~Zhang, E.~Coumans, A.~Iscen, Y.~Bai, D.~Hafner, S.~Bohez, and
  V.~Vanhoucke, ``Sim-to-real: Learning agile locomotion for quadruped
  robots,'' \emph{arXiv preprint arXiv:1804.10332}, 2018.

\bibitem{haarnoja2018learning}
T.~Haarnoja, S.~Ha, A.~Zhou, J.~Tan, G.~Tucker, and S.~Levine, ``Learning to
  walk via deep reinforcement learning,'' \emph{arXiv preprint
  arXiv:1812.11103}, 2018.

\bibitem{yang2020data}
Y.~Yang, K.~Caluwaerts, A.~Iscen, T.~Zhang, J.~Tan, and V.~Sindhwani, ``Data
  efficient reinforcement learning for legged robots,'' in \emph{Conference on
  Robot Learning}.\hskip 1em plus 0.5em minus 0.4em\relax PMLR, 2020, pp.
  1--10.

\bibitem{ha2020learning}
S.~Ha, P.~Xu, Z.~Tan, S.~Levine, and J.~Tan, ``Learning to walk in the real
  world with minimal human effort,'' \emph{arXiv preprint arXiv:2002.08550},
  2020.

\bibitem{peng2020dog}
X.~B. Peng, E.~Coumans, T.~Zhang, T.-W. Lee, J.~Tan, and S.~Levine, ``Learning
  agile robotic locomotion skills by imitating animals,'' \emph{arXiv preprint
  arXiv:2004.00784}, 2020.

\bibitem{hutter2017anymal}
M.~Hutter, C.~Gehring, A.~Lauber, F.~Gunther, C.~D. Bellicoso, V.~Tsounis,
  P.~Fankhauser, R.~Diethelm, S.~Bachmann, M.~Bl{\"o}sch \emph{et~al.},
  ``Anymal-toward legged robots for harsh environments,'' \emph{Advanced
  Robotics}, vol.~31, no.~17, pp. 918--931, 2017.

\bibitem{hwangbo2018raisim}
J.~Hwangbo, J.~Lee, and M.~Hutter, ``Per-contact iteration method for solving
  contact dynamics,'' \emph{IEEE Robotics and Automation Letters}, vol.~3,
  no.~2, pp. 895--902, 2018.

\bibitem{peng2018deepmimic}
X.~B. Peng, P.~Abbeel, S.~Levine, and M.~van~de Panne, ``Deepmimic:
  Example-guided deep reinforcement learning of physics-based character
  skills,'' \emph{ACM Transactions on Graphics (TOG)}, vol.~37, no.~4, pp.
  1--14, 2018.

\bibitem{valassakis2020force_injection}
E.~Valassakis, Z.~Ding, and E.~Johns, ``Crossing the gap: A deep dive into
  zero-shot sim-to-real transfer for dynamics,'' \emph{arXiv preprint
  arXiv:2008.06686}, 2020.

\bibitem{schulman2017ppo}
J.~Schulman, F.~Wolski, P.~Dhariwal, A.~Radford, and O.~Klimov, ``Proximal
  policy optimization algorithms,'' \emph{arXiv preprint arXiv:1707.06347},
  2017.

\bibitem{bengio2009curriculum}
Y.~Bengio, J.~Louradour, R.~Collobert, and J.~Weston, ``Curriculum learning,''
  in \emph{Proceedings of the 26th annual international conference on machine
  learning}, 2009, pp. 41--48.

\bibitem{xie2020allsteps}
Z.~Xie, H.~Y. Ling, N.~H. Kim, and M.~van~de Panne, ``Allsteps:
  Curriculum-driven learning of stepping stone skills,'' \emph{arXiv preprint
  arXiv:2005.04323}, 2020.

\end{thebibliography}

\end{document}